\pdfoutput=1

\documentclass[11pt]{article}

\usepackage[final]{ACL2023}

\usepackage{times}
\usepackage{latexsym}

\usepackage[T1]{fontenc}

\usepackage[utf8]{inputenc}

\usepackage{microtype}

\usepackage{inconsolata}
\usepackage{amsmath}
\usepackage{amssymb}
\usepackage{geometry}
\usepackage{booktabs}
\usepackage{hyperref}
\usepackage{algorithm}
\usepackage{algpseudocode}
\usepackage{latexsym}
\usepackage{lineno}
\usepackage[utf8]{inputenc}
\usepackage{booktabs}
\usepackage{siunitx}
\usepackage[table]{xcolor}
\usepackage{graphicx}
\usepackage{tikz}
\usepackage{todonotes}
\usepackage{multirow}
\usepackage{tabularx}
\usepackage{mathtools}
\usepackage{txfonts}
\usepackage{siunitx}
\usepackage{xcolor}
\usepackage{graphicx}
\usepackage{booktabs}
\usepackage{multirow}
\usepackage{pgf} 
\newcommand\langrank{\textcolor{black}{\textsc{LangRank}}}
\newcommand{\uriel}[1]{\textcolor{black}{\textsc{URIEL#1}}}

\newlength\cellwidth
\newlength\cellheight
\definecolor{darkgreen}{rgb}{0.0, 0.0, 0.75}
\definecolor{lightgreen}{rgb}{0.0, 0.0, 1.0}
\definecolor{darkred}{rgb}{0.75, 0.0, 0.0}
\definecolor{lightred}{rgb}{1.0, 0.0, 0.0}
\definecolor{myyellow}{HTML}{F28522}
\definecolor{darkblue}{rgb}{0.0, 0.0, 0.75}

\newcommand{\smallerse}[1]{{\scriptsize #1}}

\newcommand{\baselinecell}[2]{%
  \begingroup
  \bfseries
  \sisetup{round-mode=places, round-precision=1, table-format=+-1.3}%
  \num{#1}\smallerse{ $\pm$ \num{#2}}%
  \endgroup
}

\newcommand{\celldata}[4]{%
  \begingroup
  \sisetup{detect-weight, round-mode=places, round-precision=1, table-format=+-1.3}

  \pgfmathparse{#4 < 0.05 ? 1 : 0}\let\isbold\pgfmathresult

  \pgfmathparse{max(-50, min(50, #3))}\let\clampedpct\pgfmathresult
  \pgfmathparse{(\clampedpct + 50) / 100}\let\ratio\pgfmathresult
  \pgfmathparse{\ratio < 0.5 ? 2*\ratio : 1}\let\redval\pgfmathresult
  \pgfmathparse{\ratio < 0.5 ? 2*\ratio : 2*(1-\ratio)}\let\greenval\pgfmathresult
  \pgfmathparse{\ratio > 0.5 ? 2*(1-\ratio) : 1}\let\blueval\pgfmathresult

  \pgfmathparse{abs(#3) > 30 ? 1 : 0}\let\usewhite\pgfmathresult

  \colorbox[rgb]{\redval,\greenval,\blueval}{%
    \makebox[\cellwidth][c]{%
      \ifdim\usewhite pt>0pt \color{white}\fi
      \ifdim\isbold pt>0pt \bfseries\fi
      \num[explicit-sign=+]{#1}\,{\scriptsize $\pm$ \num{#2}}%
    }%
  }%
  \endgroup
}

\usepackage{fancyhdr}

\title{Modality Matching Matters: Calibrating Language Distances for Cross-Lingual Transfer in \uriel{+}}

\author{
 \textbf{York Hay Ng$^{\varheartsuit*}$},\
 \textbf{Aditya Khan$^{\varheartsuit*}$},\
 \textbf{Xiang Lu$^{\clubsuit*}$},\
 \textbf{Matteo Salloum$^{\vardiamondsuit}$},\
 \textbf{Michael Zhou$^{\spadesuit}$},\\
 \textbf{Phuong Hanh Hoang$^{\varheartsuit}$},\
 \textbf{A. Seza Doğruöz$^{\blacktriangle}$},\
 \textbf{En-Shiun Annie Lee$^{\varheartsuit\hspace{1pt}\blacksquare}$}
\\
 $^{\varheartsuit}$University of Toronto, Canada\quad
 $^{\clubsuit}$University of Michigan, USA\\
 $^{\vardiamondsuit}$Harvard University, USA\quad
 $^{\spadesuit}$Carnegie Mellon University, USA\\
 $^{\blacktriangle}$LT3, IDLab, Universiteit Gent, Belgium\quad
 $^{\blacksquare}$Ontario Tech University, Canada
 \\
 \texttt{yorkng@cs.toronto.edu, adityakhan@cs.toronto.edu, jameslx@umich.edu}
}

\begin{document}
\maketitle
\thispagestyle{fancy}
\begingroup
\renewcommand\thefootnote{\fnsymbol{footnote}}
\footnotetext[1]{The authors contributed equally.}
\endgroup
\begin{abstract}

Existing linguistic knowledge bases such as \uriel{+} provide valuable geographic, genetic and typological distances for cross-lingual transfer but suffer from two key limitations. First, their one-size-fits-all vector representations are ill-suited to the diverse structures of linguistic data. Second, they lack a principled method for aggregating these signals into a single, comprehensive score. In this paper, we address these gaps by introducing a framework for type-matched language distances. We propose novel, structure-aware representations for each distance type: speaker-weighted distributions for geography, hyperbolic embeddings for genealogy, and a latent variables model for typology. 
We unify these signals into a robust, task-agnostic composite distance. Across multiple zero-shot transfer benchmarks, we demonstrate that our representations significantly improve transfer performance when the distance type is relevant to the task, while our composite distance yields gains in most tasks.

\end{abstract}

\section{Introduction}\label{sec:introduction}
\begin{figure*}[t!]
    \centering
    \includegraphics[width=1\linewidth]{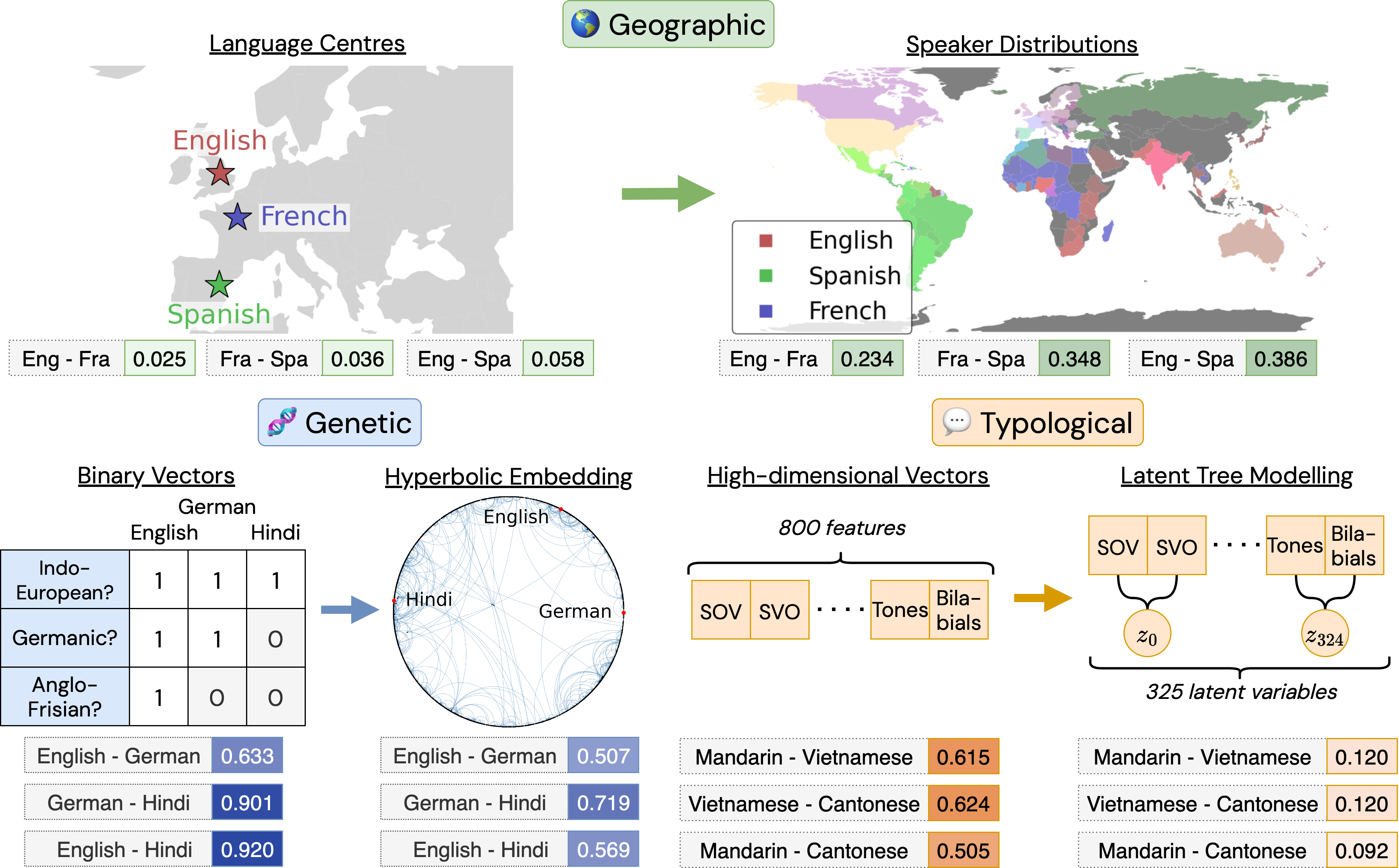}
    \caption{A demonstration of URIEL+ language representations versus our proposed representations, for each modality. Distance scores are shown for \uriel{+} (left number) and our proposed representation (right number). Lower values indicate greater similarity. Our proposed distances encode structural similarity in their respective modalities, rather than literal phylogenetic, typological, or geographic distance.}
    \label{fig:method}
\end{figure*}

Linguistic knowledge bases such as \uriel{}/\uriel{+} \cite{littell-etal-2017-uriel, khan-etal-2025-uriel} are foundational tools that quantify linguistic distance for over $7,000$ languages. These distances fall into three \textit{modalities}, or feature categories: geographic (locations of languages), genetic (linguistic family trees),  and typological (linguistic features unique to each language)\footnote{The typological modality is also commonly referred to as featural (e.g. in \citealp{khan-etal-2025-uriel}).}, as shown in Figure \ref{fig:method}. These measures are widely used in cross-lingual transfer research to assess and leverage linguistic similarity between languages for tasks such as selecting source languages for model training \citep{lin-etal-2019-choosing, lauscher-etal-2020-zero, ruder-etal-2021-xtreme, blaschke-etal-2025-analyzing, de-vries-etal-2022-make}.

As indicated by \citet{toossi-etal-2024-reproducibility}, \uriel{} represents languages in all three modalities as high-dimensional Euclidean vectors, compared via angular distance.  Despite enhancing data coverage and addressing usability issues, \uriel{+} \cite{khan-etal-2025-uriel} adopts the same language representation. This uniform approach is convenient but ill-suited for the diverse structures of linguistic data. That is to say, it produces less meaningful distances and limits the effectiveness of cross-lingual transfer where accurate representations of linguistic distance are paramount. In our study, we address this issue by proposing modality-specific distances from new language representations.

\paragraph{Limitations in \uriel{+} Representations}
\paragraph{Geographic} 
Both \uriel{} and \uriel{+} represent each language by a single Glottolog coordinate, with geographic vectors computed as great-circle distances to 299 fixed reference points.
This single-point proxy misses multi-country and diaspora populations. It also reflects historical or administrative geographical locations rather than current speaker distributions which is a key determinant for language contact \cite{nichols1992linguistic}. For example, English, French, and Spanish are pinned near cities such as London, Paris, and Madrid, although most speakers of these languages reside elsewhere (Figure \ref{fig:method}, Geographic). 
This can result in counter-intuitive discrepancies, causing languages with large, overlapping speaker communities to appear geographically distant and providing misleading signals for transfer.

\paragraph{Genetic} The current genetic representation flattens the Glottolog tree into sparse, one-hot vectors indicating language family membership (>3700 dimensions, 99.85\% zeros), losing the crucial hierarchical structure of genetic relationships. This flat representation counts shared ancestry at all levels equally. For example, the close relationship between German and English (Germanic) is given the same weight as the far more distant relationship between German and Hindi (Indo-European) (Figure~\ref{fig:method}, Genetic), obscuring fine-grained distinctions in genetic structure relevant for transfer.

Moreover, this representation is limited to terminal nodes (languages), failing to provide embeddings for internal nodes (language families and sub-families). Thus, it does not provide a continuous, low-dimensional representation over the genealogical structure itself.

\paragraph{Typological} High-dimensional binary feature vectors are sparse, with correlated and sometimes redundant features, weakening the ability of angular distances to capture meaningful structural similarity. For instance, features for ``Subject-Object-Verb'' and ``Subject-Verb-Object'' word order are highly correlated yet treated as independent signals, inflating distances between languages which differ on related features.~\citet{ng-etal-2025-less} empirically showed that such redundancy and high dimensionality reduce the effectiveness of typological vectors in capturing meaningful structural similarity.

Given the limitations in language representations in \uriel{} and \uriel{+} (especially for cross-lingual transfer), what makes a good language distance for transfer? We claim that each modality should use a representation and distance suited to its structure. Therefore, we embed the original \uriel{+} vectors into a representation that captures the inherent structure (e.g., the hierarchical genealogy) of each modality and compute distances on this new representation.

Another fundamental limitation of \uriel{+} is that it cannot compute a cumulative distance using all modalities. This forces researchers to choose between signals (e.g., typology or genetics), even though a unified metric is often preferred for practical applications such as transfer language selection \cite{ahuja-etal-2022-multi, srinivasan2021predictingperformancemultilingualnlp}. We address this gap by developing a composite distance: a weighted average of distances from individual modalities, providing a single value that simplifies applications in cross-lingual transfer.

Our paper rectifies the aforementioned issues with the following contributions: 
\begin{enumerate}    
    \item We formalize modality-matched language distances, introducing new representations and distance metrics for each modality.
    \begin{itemize}
        \item \textbf{Geographic} We model each language as a distribution over speaker locations instead of a single coordinate.
        \item \textbf{Genetic} We embed the Glottolog \cite{Glottolog} family tree in hyperbolic space, producing a low-dimensional hierarchical representation.
        \item \textbf{Typological} We group correlated features into latent variables (“islands”), producing a compact representation that captures structural patterns.
    \end{itemize}
    \item We propose a simple composite distance that aggregates modality-specific distances. 
\end{enumerate}

Empirically, across cross-lingual transfer benchmarks with \langrank{} \cite{lin-etal-2019-choosing}, modality-matched distances consistently improve source language selection.

\paragraph{Key Findings}
\begin{enumerate}
\item Language representations aligned with the latent structure of each modality leads to statistically significant improvements in transfer language selection compared to \uriel{+} \cite{khan-etal-2025-uriel}.
\item In transfer performance, the impact of any single modality is task-dependent, confirming and extending \citet{blaschke-etal-2025-analyzing}: transfer performance is sensitive not only to the distance measure(s) used, but also to the choice of language representations.
\item Aggregating modality-matched distances into a composite score yields a single, task-agnostic measure that often outperforms \uriel{+} even without task-specific training.
\end{enumerate}

\section{Related Research}\label{sec:related-work}
\paragraph{\uriel{} in Cross-Lingual Transfer}
\uriel{} distances serve as a strong predictor of transfer performance \cite{khiu-etal-2024-predicting, philippy-etal-2023-identifying, lauscher-etal-2020-zero, tran-bisazza-2019-zero} between languages, performing comparably to other linguistic measures \cite{Eronen_2023}.

Consequently, \uriel{} distances have been widely applied to enhance cross-lingual transfer, particularly in predicting the performance of multilingual models \cite{anugraha-etal-2025-proxylm, srinivasan2021predictingperformancemultilingualnlp, xia-etal-2020-predicting, patankar-etal-2022-train}, selecting transfer languages \cite{lin-etal-2019-choosing, Eronen_2023}, and language model regularization \cite{adilazuarda-etal-2024-lingualchemy}, demonstrating its indispensable role in multilingual natural language processing (NLP).

\paragraph{Distributional Representation of Geographic Data} 

Moving from "language as a point" to "language as a distribution" is crucial for capturing signals from language contact \citep{dunn-edwards-brown-2024-geographically, nichols1992linguistic}. Empirical audits show that single-point geography can mask biases in data by under-representing where speakers actually reside \citep{faisal-etal-2022-dataset}. A natural method for comparing speaker distributions is the Wasserstein-1 distance (or Earth Mover's distance) \citep{Villani2009}, which measures the minimum ``work'' needed to transform one distribution into another. Optimal transport has proven effective in NLP for tasks such as measuring document similarity \citep{pmlr-v37-kusnerb15}, evaluating text generation \citep{clark-etal-2019-sentence}, and aligning word embeddings \citep{zhang-etal-2017-earth}, making it a well-grounded choice for our geographic modality.

\paragraph{Sparser Representations of Typological Data} 

Typological feature sets are often high-dimensional, redundant, and noisy \cite{ng-etal-2025-less}, with inconsistent feature choices yielding wide variation across studies  \citep{ploeger-etal-2024-typological,poelman-etal-2024-call}. Compact, structured representations can mitigate these issues, improving typology-driven downstream tasks such as machine translation, cross-lingual evaluation, and data or language selection \citep{Bjerva-typology,ploeger2025,hlavnova-ruder-2023-empowering,adilazuarda-etal-2024-lingualchemy, brinkmann-etal-2025-large}.

To achieve this, we turn to latent tree models (LTMs), which can uncover hidden structure from data without supervision. By grouping correlated features and capturing unobserved confounders, LTMs produce task-agnostic, denoised embeddings \citep{zwiernik2017,williams2018} that have proven effective for related tasks such as topic discovery and sentence modeling \citep{Mourad_2013,CHEN2017105,williams2018}.

\paragraph{Hyperbolic Representations of Genetic Data}
Euclidean space (with flat curvature and polynomial volume growth) poorly fits data where latent structure is tree-like, and leads to unnecessary distortion. \uriel{+} vectors lie in such a flat space (see Appendix \ref{app:genetic}). Instead, hyperbolic geometry offers a closer match as its exponential volume growth aligns with the branching of trees, enabling low-distortion, low-dimensional embeddings. \citet{NickelKiela2017} showed that Poincar\'e-ball embeddings capture WordNet hierarchies with markedly less distortion and in fewer dimensions than Euclidean baselines. Extending this idea, \citet{tifrea2018poincareglovehyperbolicword} adapted the commonly used GloVe model to learn directly in hyperbolic space, improving word similarity, analogy, and especially hypernymy detection. Beyond the Poincar\'e model, the hyperboloid (Lorentz) model embeds points in Minkowski space, simplifying certain operations and often improving numerical stability during training \citep{NickelKiela2018}.

In multilingual NLP, incorporating linguistic genealogy assists cross-lingual transfer (e.g., by guiding meta-learning with genetic structure or by arranging adapter modules to mirror the language tree \citep{garcia-etal-2021-cross,faisal-anastasopoulos-2022-phylogeny}). Prior hyperbolic work on languages used cognate similarity to infer hierarchical relations \citep{NickelKiela2018}.

To the best of our knowledge, our work is the first to directly embed the comprehensive language hierarchy from Glottolog \cite{Glottolog} to hyperbolic space, providing a novel application and a rigorous empirical comparison of foundational geometric embedding techniques on this linguistic resource.

\begin{table*}[t]
\centering
\small
\setlength{\tabcolsep}{4pt}
\renewcommand{\arraystretch}{1.12}
\begin{tabular}{
  >{\raggedright\arraybackslash}p{2.0cm}
  >{\raggedright\arraybackslash}p{3.2cm}
  >{\raggedright\arraybackslash}p{3.2cm}
  >{\raggedright\arraybackslash}p{3.5cm}
  >{\raggedright\arraybackslash}p{3.1cm}
}
\toprule
\textbf{Modality} & $\mathcal X^m$ & $\mathcal Z^m$ & $f^m$ & $d^m$ \\
\midrule
Geography &
Country speaker counts + centroids &
Distribution over locations (speaker shares) &
Normalise counts to a probability distribution &
Earth Mover’s distance \\
Genetic &
Glottolog genealogy &
Hyperboloid Embeddings &
Learn embeddings &
Hyperbolic distance \\
Typology &
Binary features &
Posteriors over latent ``islands'' &
Fit islands; map to posteriors &
Angular distance \\
\bottomrule
\end{tabular}
\caption{Summary of modality representations and their distances. Distances are normalized and may be aggregated into a composite distance.}
\label{tab:modality-summary}
\end{table*}

\paragraph{Need for a Composite Distance Score}
A recurring challenge in cross-lingual work is the need to juggle multiple, often task-dependent, linguistic distances without a single, reusable score. While resources such as ~\citet{khan-etal-2025-uriel} provide individual distances, they do not offer a principled way to aggregate them. Some methods fuse modalities within a training objective (e.g., \textsc{LinguAlchemy} regularises with typological, geographic, and genetic vectors), but these do not yield a calibrated, standalone language-to-language distance metric \citep{adilazuarda-etal-2024-lingualchemy}. This motivates our goal of creating a single, normalized composite score usable across tasks and languages.

\paragraph{Representation Requirements From Prior Work}
Synthesizing the evidence above, we adopt four requirements for cross-lingual distance:
\begin{itemize}
  \item \textbf{Geography as distributions}: Languages should be represented as dispersed speaker distributions, not as single points.
  \item \textbf{Genealogy as hierarchy}: Distances should respect language ancestor–descendant structure.
  \item \textbf{Typology as low-noise factors}: Redundant/correlated features should be compressed into a compact representation.
  \item \textbf{Composability}: Modality-specific distances should be normalized so they can be aggregated into a single composite score.
\end{itemize}

\section{Modality Representations and Cross-Modal Composition}
The central premise in this work is that each modality benefits from a representation that matches its latent structure. To illustrate, we briefly review the modalities in \uriel{+}, and introduce our modality matched representations and distances along with describing we may combine them. A summary of the representations is presented in Table \ref{tab:modality-summary}. 

\subsection{Formalizing Modalities}
Let $\mathcal{L}$ denote the set of languages and let $M$ denote modalities in \uriel{+}: $$M = \{\text{geography},\ \text{genetic},\ \text{typology}\}.$$ For each modality $m\in M$, let $\mathcal{X}^m$ be the raw data space (e.g. country/territory speaker counts for geography, the Glottolog genealogy counts for genetic, binary typology vectors). For a language $\ell\in\mathcal{L}$, we write $x_\ell(m)\in\mathcal{X}^m$ for its raw modality-specific data. For example, $x_\text{German}(\text{geo})$ corresponds to the geography vector for the German language in \uriel{+}. 

For each $m\in M$ we specify a representation mapping $f^m: \mathcal{X}^m \to \mathcal{Z}^m$, where $\mathcal{Z}^m$ is an appropriate representation space. For instance, if $m =$ genetic, then $\mathcal{Z}^m$ has to capture the hierarchical structure of the family tree of a particular language. After representing each modality vector for a language $\ell$ in the new representation space, denoted $f^m(x_\ell(m))$, we compute distances between these using a normalized distance $d^m \in [0, 1]$ defined on $\mathcal{Z}^m$. 

\subsection{Geography as Distributions}\label{subsec:geography}
Representing a language with a single point ignores effects from language contact, arising from multi-country speaker populations shaped by globalization and migration. Contrarily, modeling languages by the geographical distribution of speakers captures dispersion and overlap across regions. By comparing the distance between speaker distributions, we obtain a population-aware geographic signal that better reflects the geographic proximity of languages.

We source from Ethnologue \cite{eberhard2025ethnologue28} the number of language speakers per language per country to model each language as a discrete probability distribution over locations, with mass proportional to the share of speakers at those locations. We use the total speaker count from Ethnologue, due to its broad language coverage and standardized data collection. However, we acknowledge that this choice presents reproducibility challenges (see Limitations). In particular, for language $\ell \in \mathcal L$, let the location (i.e. countries or territories) where $\ell$ is spoken be indexed by $i=1,\dots,r$, with geographic centroids $y_i\in\mathbb{S}^2$ (WGS84) and speaker counts $n_{\ell,i}\ge 0$ \cite{karney-2013-geodesics}.
To calculate the distance between these speaker distributions, we normalize speaker counts $n_{\ell, i}$ in each location $i$, yielding the share of speakers of language $\ell$ at location $i$, $q_{i}$. This produces the distribution $\mathbb{P}_\ell = \{(y_i, q_{i})\}_{i=1}^r.$  Essentially, each language $\ell$ is represented by a list of locations (represented as coordinates) with weight $q_{i}$ corresponding to the proportion of the language's speakers residing there. For languages attested in only a single country, we represent the language by its Glottolog coordinate instead to preserve the granular information provided by Glottolog. We therefore define $f^{\text{geo}}$ as the mapping $x_\ell(\text{geo}) \mapsto \mathbb{P}_\ell$.

A natural distance measure $d^{\text{geo}}$ between speaker distributions is the Earth Mover distance \cite{Villani2009}. To define this, suppose that $\ell_1 \mapsto \mathbb{P}_{\ell_1} = \{(y_i, q_{i})\}_{i=1}^r$ and $\ell_2 \mapsto \mathbb{P}_{\ell_2} = \{(z_i, v_{i})\}_{i=1}^n$. We define the set of feasible transport plans 
\[
\Pi(\mathbb{P}_{\ell_1}, \mathbb{P}_{\ell_2})
= \left\{ \pi\in\mathbb{R}_{\ge0}^{r\times n}\ \middle|\ 
\substack{\sum_j \pi_{ij}=q_i\\[2pt] \sum_i \pi_{ij}=v_j}
\right\}
\]
Allowing us to define language distance as $$d^{\text{geo}}(\ell_1, \ell_2) = \frac{1}{D_{\max}} \min_{\pi \in \Pi} \sum_{i=1}^r \sum_{j=1}^n \pi_{ij} d_g(y_i, z_j)$$ where $d_g$ is the shortest distance between the two geographic centroids that remain on the Earth's surface, also known as the geodesic distance; and $D_{\max} = \max_{x,y\in\mathbb{S}^2} d_g(x,y)$, representing the geodesic distance between the two poles on Earth. This metric iterates through all possible methods of transforming one speaker distribution into another, choosing the one requiring the least work. Normalization then yields a distance between speaker distributions. 
A proof that this normalization yields values in $[0,1]$ is provided in Appendix \ref{app:geographic}.

\subsection{Genealogy as Hierarchy}
To overcome the issues described in Section \ref{sec:introduction}, we propose a principled, structure-preserving approach by learning dense embedding vectors for the entire Glottolog genealogical tree, including families, languages, and optionally dialects, in a low-dimensional, continuous space. The ideal geometric space for this task is hyperbolic geometry, whose metric properties are intrinsically suited for representing hierarchical data with minimal distortion. The space's negative curvature and exponential volume growth provide a natural geometric analogue to the branching, tree-like structure of linguistic evolution, where the number of descendants grows exponentially with depth from the proto-language root. This hyperbolic approach, while not intended to redefine phylogenetic relatedness, aims to encode the genealogical structure of Glottolog in a geometry suitable for downstream modeling.

Formally, we represent the Glottolog genealogy tree as a directed acyclic graph $G=(V,E)$, where $V$ is the set of linguistic entities (nodes), and $E$ contains the directed parent-to-child edges. Our goal is to learn an embedding function $f^{\text{gen}} : V \to \mathcal{H}^d$ that maps each node $v \in V$ to a point in the $d$-dimensional hyperbolic space. We explored two isometric models of hyperbolic geometry: the Poincar\'e disk model and the hyperboloid model, and denote the hyperbolic distance between $a$ and $b$ as $d_{\text{Hyp}}(a, b)$.
The learning objective is designed to encourage the geometric arrangement of embeddings in $\mathcal{H}^d$ to faithfully reflect the complete genealogical topology of $G$. To enforce this globally, we define our set of positive training pairs, $\mathcal{P}$, as the transitive closure of the parent-child edges in $E$, meaning that a pair $(u, v) \in \mathcal{P}$ if and only if $u$ is an ancestor of $v$. Hence, following \citet{NickelKiela2017, NickelKiela2018}, for each positive pair $(u,v) \in \mathcal{P}$, we adopt a contrastive objective, sampling $K$ negative nodes $\{w_1, \dots, w_K\}$ that are not descendants of $u$, and define the objective per pair as $$\text{L}_{(u,v)} = - \log \frac{\exp(-d(u,v))}{\exp(-d(u,v)) + \sum_{i=1}^K \exp(-d(u,w_i)) }.$$

The total objective is $\text{L}_{(u,v)}$ summed over all positive pairs: $\text{L} =  \sum_{(x,y) \in \mathcal{P}} \text{L}_{(x, y)}$. Maximizing this objective pulls each positive pair closer to each other while simultaneously pushing negative pairs farther apart, thus encouraging hierarchical fidelity.

The derived distance metric on $\mathcal Z^m$ is given by $d^{\text{gen}} =  d_{\text{Hyp}}(a, b) / D_{\max}$. Here $D_{\max}$ is the maximum pairwise hyperbolic distance. This ensures that the distance is bounded in $[0, 1]$. In preliminary experiments, the hyperboloid model performed stronger in ancestor retrieval tasks. Thus, we adopt the hyperboloid embeddings and distance metric for \langrank{} experiments and evaluation. Further details are in Appendix \ref{app:genetic}.

\begin{table*}[h!]
\centering
\begin{tabularx}{\textwidth}{l c X c c c c} 
\toprule
\textbf{Task Type} & \textbf{Dataset} & \textbf{Related Work} & \textbf{Model} & \textbf{Metric} & \textbf{Target} & \textbf{Source} \\
\midrule
Machine Trans. & TED & \citet{lin-etal-2019-choosing} & RNN+Attn & BLEU & 54 & 54 \\
\multirow{2}{*}{Dep. Parsing} & UD v2.2 & \citet{lin-etal-2019-choosing} & Biaffine & Accuracy & 30 & 30 \\
& UD v2.14 & \citet{blaschke-etal-2025-analyzing} & UDPipe 2 & LAS & 152 & 70 \\
\multirow{2}{*}{POS Tagging} & UD v2.2 & \citet{lin-etal-2019-choosing} & BiLSTM & Accuracy & 60 & 26 \\
& UD v2.14 & \citet{blaschke-etal-2025-analyzing} & UDPipe 2 & UPOS & 152 & 70 \\
Entity Linking & Wikipedia & \citet{lin-etal-2019-choosing} & BiLSTM & Accuracy & 54 & 9 \\
\multirow{2}{*}{Topic Class.} & Taxi1500 & -- & mBERT\footnotemark[2] & Macro F1 & 799 & 33 \\
 & SIB200 & \citet{blaschke-etal-2025-analyzing} & XLM-R & Macro F1 & 197 & 160 \\
NLI & XNLI & \citet{philippy-etal-2023-identifying} & mBERT & Accuracy & 15 & 15 \\
\bottomrule
\end{tabularx}
\caption{List of the NLP tasks applied to \langrank. ``Target'' and ``Source'' refers to the number of source and target languages where models are tested and trained on, respectively. Related works link to previous applications in choosing transfer languages based on language distances.  }
\label{tab:langrank_tasks}
\end{table*}

\subsection{Typology as Low-Noise Factors}
A natural choice to model confounding variables and inherent structure in language typology is latent tree models (LTM). We use this to cluster typological features into groups (termed ``islands'' and denoted as $G_i$) governed by latent variables that capture confounding variables, co-occurrence structure, while addressing redundancy. We obtain a dimensionality reduction mapping $f^{\text{typ}}$ from this method. 

Given a subset of binary typological features $t_\ell=(t_{\ell,1},\dots,t_{\ell,s})$,
we introduce a binary latent variable $z_i\in\{0,1\}$ for island $i$ and parameters
\[
\theta^{(i)}_{jk} := \mathbb{P}(t_{\ell,j}=1 \mid z_i=k),
\quad j\in G_i,\ k\in\{0,1\}.
\]
learned by Expectation–Maximization \cite{Dempster1977EM}, where priors are initialized uniformly and conditionals are initialized randomly. We perform early stopping via a modified Bayesian Information Criterion (BIC) \footnote{See Appendix \ref{app:latent} for implementation details.} which penalizes log-likelihood and the number of parameters quadratically, encouraging more balanced clusters.

To scale beyond a single latent variable, we implement a greedy algorithm to obtain multiple ``islands''. Iteratively, we repeat the following process: (i) initialize an active set using the pair of features with highest Mutual Information (MI) \cite{Peng2005mRMR} not yet assigned to any latent variable; (ii) add the feature yielding the highest MI with the features in the active set; (iii) attempt to split the active set into two using the modified BIC; (iv) if the split is preferred, refine by testing feature switches across the two groups to further improve BIC. When a split is accepted, we obtain two groups $G_1,G_2$. We define the larger group as an island, associating it with a latent variable $z_i$, and store its $s_i\times 2$ parameter matrix $(\theta_{jk}^{(i)})$ as a cluster. Here, $z_i$ is the latent variable for the $i$th island, and $s_i$ is the number of features assigned to island $i$. The remaining features return to the pool and the process repeats.

Finally, a typological vector $x_\ell(\text{typ})$ is mapped to the concatenated posterior vector
\[
\mathbf{p}(t_\ell)
:= \left(\mathbb{P}(z_i=0\mid t_{\ell,G_i}),\ \mathbb{P}(z_i=1\mid t_{\ell,G_i})\right)_{i=1}^n{}^\top .
\]
where $t_{\ell,G_i}$ denotes the subvector of $t_\ell$ restricted to the features in island $G_i$, and $n$ is the number of islands. This representation is naturally normalized per island. We compute angular distances on our representation, as is done by default in \citet{khan-etal-2025-uriel}, due to its sensitivity to the proportional relationships between posterior probabilities across islands, rather than their absolute magnitudes; thus making it a robust metric for comparing the structural profiles of languages.

\subsection{Composability: Aggregating Distances}
\label{subsec:composite}
Practitioners often desire a single distance score between languages. Given nonnegative modality weights $w\in\mathbb{R}_{\ge 0}^{|M|}$ with $\sum_{m\in M} w_m=1$, we define the normalized composite distance
$$
D(\ell_i,\ell_j):=\sum_{m\in M} w_m d^m\left(f^m(x_{\ell_i}(m), x_{\ell_j}(m))\right).
$$

Although the weights can be learned specifically for a given cross-lingual transfer task, the simplest case is to simply let $w_m = 1/|M|$ for all $m$. In doing so, $D$ collapses to a simple average--this serves as a strong default. It assumes the user does not favor any particular modality a priori when evaluating how distant language is. Furthermore, it is simple and robust, requiring no task-specific tuning. Nonetheless, we present alternative ways to select weights in Appendix \ref{app:composite-weights}.

\begin{table*}[!t]
\centering
\begingroup
\footnotesize
\setlength{\tabcolsep}{2pt}
\setlength{\fboxsep}{2pt}
\begin{tabular}{l l c c c c}
\toprule
\textbf{Modality} & \textbf{Representation} & DEP & EL & MT & POS \\
\midrule
\multicolumn{2}{c}{Baseline:}   & \baselinecell{11.4}{2.9} & \baselinecell{30.0}{6.2} & \baselinecell{12.5}{1.8} & \baselinecell{27.9}{4.4} \\
\midrule
\multirow{2}{*}{Typ} & Laplacian & \celldata{0.8}{1.0}{6.5}{0.436} & \celldata{-3.8}{2.8}{-12.5}{0.177} & \celldata{0.7}{0.9}{5.2}{0.460} & \celldata{-2.1}{1.9}{-7.3}{0.276} \\
& Islands    & \celldata{0.5}{1.0}{4.4}{0.603}   & \celldata{-1.2}{2.8}{-3.9}{0.670}  & \celldata{-1.0}{0.9}{-8.3}{0.237}       & \celldata{-0.4}{1.9}{-1.4}{0.838}   \\
Geo & Speaker & \celldata{0.6}{0.7}{5.6}{0.347}  & \celldata{-7.4}{2.0}{-24.6}{0.0} & \celldata{-1.0}{0.6}{-7.6}{0.128}    & \celldata{-0.3}{1.3}{-1.0}{0.839}    \\
Gen & Hyperbolic    & \celldata{-0.9}{0.7}{-7.8}{0.192}  & \celldata{3.6}{2.0}{11.9}{0.069}    & \celldata{-4.5}{0.6}{-36.1}{0.0} & \celldata{-1.0}{1.3}{-3.5}{0.467}    \\
\bottomrule
\end{tabular}
\endgroup

\vspace{1em}

\begingroup
\footnotesize
\setlength{\tabcolsep}{2pt}
\setlength{\fboxsep}{2pt}
\begin{tabular}{l l c c c c c}
\toprule
\textbf{Modality} & \textbf{Representation} & Taxi1500 & SIB200 & XNLI & UD2.14 POS & UD2.14 DEP \\
\midrule
\multicolumn{2}{c}{Baseline:}   & \baselinecell{38.1}{0.5} & \baselinecell{16.9}{1.1} & \baselinecell{6.2}{1.2} & \baselinecell{27.4}{1.5}   & \baselinecell{35.6}{1.9}    \\
\midrule
\multirow{2}{*}{Typ} & Laplacian        & \celldata{0.4}{0.3}{1.0}{0.216} & \celldata{-0.2}{0.5}{-1.3}{0.630}    & \celldata{0.4}{0.6}{6.9}{0.465}   & \celldata{1.8}{0.8}{6.7}{0.025}     & \celldata{1.5}{0.9}{4.1}{0.103}    \\
& Islands    & \celldata{-0.9}{0.3}{-2.5}{0.003}     & \celldata{-1.4}{0.5}{-8.3}{0.003} & \celldata{-2.4}{0.6}{-38.1}{0.000}  & \celldata{-0.6}{0.8}{-2.2}{0.471}  & \celldata{-1.8}{0.9}{-5.0}{0.049}   \\
Geo & Speaker & \celldata{-2.1}{0.2}{-5.6}{0.0}   & \celldata{-0.6}{0.3}{-3.5}{0.071} & \celldata{0.1}{0.4}{2.0}{0.760}    & \celldata{-1.6}{0.6}{-5.8}{0.006}  & \celldata{0.7}{0.6}{1.9}{0.295}   \\
Gen & Hyperbolic    & \celldata{2.7}{0.2}{7.0}{0.0}     & \celldata{1.0}{0.3}{6.0}{0.002} & \celldata{-0.1}{0.4}{-1.3}{0.839}      & \celldata{-2.6}{0.6}{-9.5}{0.0}  & \celldata{-3.9}{0.6}{-11.0}{0.0}   \\
\bottomrule
\end{tabular}
\endgroup

\caption{The impact of distance metrics on performance loss when picking the top transfer language from \langrank{}. Values are regression coefficients $\pm$ standard error, measured in percentage points. Baseline rows represent the intercept, indicating the performance loss when using \uriel{+} representations for each modality. Lower is better. Results where $p < 0.05$ are shown in \textbf{bold}. Color corresponds to the percentage change in performance loss.}
\label{tab:results}

\end{table*}

\section{Validation on Downstream Tasks}\label{sec:evaluation}

Although prior work on evaluating distance measures have mostly explored the impact of individual distances on transfer performance \cite{lauscher-etal-2020-zero, philippy-etal-2023-identifying, blaschke-etal-2025-analyzing}, we illustrate the real-world utility and isolated impact of our language representations in enhancing cross-lingual transfer by applying \langrank{} \cite{lin-etal-2019-choosing}, a widely used framework for choosing transfer (source) languages for cross-lingual NLP tasks. Given a set of language distances, \langrank{} uses gradient-boosted decision trees to select transfer languages for a given task and target language.

\subsection{Experimental Setup}
\footnotetext[2]{We additionally experiment on LLaMA-3.1-8B for Taxi1500, see Appendix \ref{app:downstream-llama}.}

Table \ref{tab:langrank_tasks} lists the tasks studied. Based on the findings in \citet{blaschke-etal-2025-analyzing}, we augment the original \langrank{} framework with five new tasks: Taxi1500 \cite{ma-etal-2025-taxi1500}, due to its substantial language coverage; XNLI \cite{conneau-etal-2018-xnli}, SIB200 \cite{adelani-etal-2024-sib}, along with dependency parsing and part-of-speech tagging tasks from Universal Dependencies \cite{nivre-etal-2020-universal}, where the relationship between transfer performance and language distance was previously determined \citep{philippy-etal-2023-identifying, blaschke-etal-2025-analyzing}. We intentionally mirror prior work in transfer language selection, including their choice of models and datasets. This expanded evaluation enables direct comparison and replication, while supporting the generalizability of our findings across tasks and languages.

We utilize ``performance loss'' to measure how well \langrank{} enhances cross-lingual performance in NLP tasks. Performance loss is defined as the relative loss in performance when transferring from the top-1 language chosen by \langrank{}, compared to the performance of the optimal source, for a given target language. \footnote{See Appendix \ref{app:downstream-metric} for the formal definition.} This setup  demonstrates the real-world impact of language representations on cross-lingual transfer more accurately.

Using only language distances as features, we conduct an ablation study by training \langrank{} with distances from different representations \footnote{See Appendix \ref{app:downstream} for the full setup, and hyperparameters.}. For the genetic modality, we ablate on the \uriel{+} and hyperbolic representations. For the typological modality, we additionally ablate on the representation applying Laplacian Score feature selection \cite{he2005laplacian} on \uriel{+} typological vectors, which was found to be a robust selection method for \langrank{} in  \citet{ng-etal-2025-less}. Within each ablation and task, we conduct leave-one-language-out cross-validation (i.e. testing performance loss for each target language, before averaging).

Collecting scores across folds and ablations, we fit a linear mixed-effects model with performance loss as the dependent variable, three categorical variables indicating the representation used as fixed effects, with the intercept measuring baseline \uriel{+} performance. An additional random intercept is placed on the cross-validation fold. Model parameters are estimated via L-BFGS optimization. This approach estimates the impact of each representation, while accounting for variability across folds. To further assess relevance to modern architectures, we additionally report results with LLaMA-3.1 on Taxi1500 in Appendix \ref{app:downstream-llama}.

\subsection{Results}
The isolated impact of our new representations on cross-lingual transfer performance is detailed in Table \ref{tab:results}. First, we observe that baseline performance losses varied from 6.2 - 38.1 between tasks, confirming that, even when applying \uriel{+} distance measures, \langrank{} remains a viable and robust choice for choosing transfer languages.

Next, there usually exists combinations of language representations that significantly improve cross-lingual performance. Notably, our modality-matched representations can substantially reduce transfer error. For example, in the XNLI task, using our latent islands representation for typology reduces the baseline performance loss of 6.2 by 2.4 points (a 39\% improvement). Similarly, for Machine Translation, our hyperbolic genetic embeddings reduce the baseline loss of 12.5 by 4.5 points (a 36\% improvement).

Crucially, when comparing datasets that instantiate the same NLP task (e.g., Taxi1500 vs. SIB200, both topic classification tasks), we observe no contradictions among statistically significant results. A representation that significantly improves transfer in one dataset never significantly degrades performance in another within the same NLP task.

These consistent reductions in performance loss highlight how our representations generally outperform \uriel{+}, in particular for the low-resource languages in our evaluation (e.g. Taxi1500 contains 764 low-resource languages\footnote{Defined as language classes 0-2 from \citet{joshi-etal-2020-state}.}). Through aligning representations and distance metrics with the inherent structure of linguistic modalities, our framework unlocks more nuanced signals for cross-lingual transfer. 

These results simultaneously illustrate a cautionary tale. Although our representations can significantly improve performance, there are instances where swapping out \uriel{+} representations worsens performance. This task-dependent variability suggests a deeper interplay between the nature of a task and the linguistic information most relevant to it. We hypothesize that tasks highly sensitive to language contact and lexical borrowing, such as certain classification or entity linking tasks, benefit most from our speaker distribution model, which explicitly captures geographic overlap. 

Conversely, tasks where syntactic structure is relevant might have a more complex relationship with genealogy. While our hyperbolic embeddings more faithfully model the Glottolog hierarchy, the transferability of syntax may be influenced more by recent, horizontal contact phenomena or areal features not captured by vertical descent alone. Overall, the finding that transfer performance depends on both the task and language representation used aligns with \citet{blaschke-etal-2025-analyzing}; therefore, we find no one-size-fits-all distance measure for cross-lingual transfer.

\begin{table}[ht!]
\centering
\setlength{\tabcolsep}{4pt}
\renewcommand{\arraystretch}{1.1}
\begin{tabular}{lccc}
\toprule
\textbf{Task}  & DEP & EL & MT \\
\textbf{Score} & 9.9 (\textcolor{darkblue}{$\downarrow 1.5$}) &
                 25.6 (\textcolor{darkblue}{$\downarrow 4.4$}) &
                 11.2 (\textcolor{darkblue}{$\downarrow 1.3$}) \\
\midrule
\textbf{Task}  & POS & XNLI & Taxi \\
\textbf{Score} & 22.8 (\textcolor{darkblue}{$\downarrow 5.1$}) &
                 3.5 (\textcolor{darkblue}{$\downarrow 2.7$}) &
                 46.7 (\textcolor{darkred}{$\uparrow 8.6$}) \\
\midrule
\textbf{Task}  & SIB & POS 2 & DEP 2 \\
\textbf{Score} & 14.4 (\textcolor{darkblue}{$\downarrow 2.5$}) &
                 21.3 (\textcolor{darkblue}{$\downarrow 6.1$}) &
                 36.7 (\textcolor{darkred}{$\uparrow 1.1$}) \\
\bottomrule
\end{tabular}
\caption{Performance loss when choosing the top-1 transfer language using the composite distance. Parentheses show the absolute change relative to the corresponding baseline intercept in Table \ref{tab:results}; \textcolor{darkblue}{$\downarrow$} indicates lower loss (better), \textcolor{darkred}{$\uparrow$} indicates higher loss (worse).}
\label{tab:composite}
\end{table}

\paragraph{Composite Distances} We additionally benchmark the performance loss incurred when choosing transfer languages based on the composite distance measure from Section \ref{subsec:composite}. Defining $w_m=\frac{1}{|M|}$, this distance measure simply averages over distances from our new representations in each modality.

The utility of this composite distance is shown in Table \ref{tab:composite}. Our results demonstrate that this composite distance serves as a strong general-purpose baseline. On most of the tasks evaluated, including Entity Linking (25.6 vs. a baseline of 30.0) and XNLI (3.5 vs. a baseline of 6.2), it reduces performance loss compared to using \uriel{+} distances alone. However, this aggregation is not uniformly optimal across all tasks, reinforcing findings from \citet{blaschke-etal-2025-analyzing, goot-etal-2025-distals}. Its substantial under-performance on tasks such as Taxi1500 classification (46.7 loss vs. a baseline of 38.1) highlights that a simple, unweighted average can obscure the most important modality for certain applications. Although the composite distance does not dominate task-specific selection models, it nevertheless offers a conservative yet robust and reusable alternative that does not necessitate task-specific training.

This metric addresses a long-standing need in the community for a single, robust score for language similarity. Additionally, our framework enables future work in learning weights based on relevance to specific tasks, which would yield supplementary performance gains and derive insights into the relevance of specific modalities to transfer performance in different NLP tasks. 

\section{Conclusion}
We presented a new framework for computing linguistic distance based on modality-matched representations. Our novel, structure-aware methods for geography (speaker distributions), genealogy (hyperbolic embeddings), and typology (latent feature islands) were designed to better capture the unique characteristics of each linguistic signal.

Our experiments confirm that the utility of these representations is fundamentally task-dependent--no single metric is optimal for all scenarios. This finding reframes our contribution as a flexible toolkit for cross-lingual research, empowering practitioners to choose the most suitable distance metric for their specific application. As a general alternative, we propose a composite distance that averages these signals. While this score provides a strong, general-purpose baseline that improves over \uriel{+} on a majority of the tasks we tested, its sub-optimal performance on some tasks highlights that aggregation trades task-specific optimality for broad applicability. To encourage community participation, we release all our code for more principled investigations into linguistic distance: \url{https://github.com/Swithord/urielplus-modality-matters}.

\section*{Limitations}

\paragraph{Data Sources}
Our work fundamentally relies on existing linguistics sources, and therefore inherits any inaccuracies or incomplete data, which may affect the quality of language representations unequally. In particular:
\begin{itemize}
\item Our speaker distribution model is founded on the basis that geographic proximity of speakers influence language contact, but this model is constrained by the granularity and scope of Ethnologue. It relies on national-level speaker counts, which may not accurately capture the precise distribution of speakers. Additionally, Ethnologue does not consider other factors influencing speaker interactions, such as time, topography, and culture. Furthermore, as the data from Ethnologue is proprietary, this prevents us from fully publicly releasing our representations.
\item Hyperbolic embeddings are designed to solely model the Glottolog tree. However, Glottolog represents only one specific model of language history that is subject to ongoing linguistic research and revision. Moreover, while we choose to embed all Glottolog languoids including dialects, we recognize that Glottolog's coverage of dialects may not be comprehensive.

\item Our latent feature islands method offers another representation of \uriel{+}'s typological data, but remains subject to the issue of sparsity. Specifically, 87\% of values in \uriel{+} are missing prior to imputation \cite{ng-etal-2025-less}. This impacts the accuracy of our representations, with potentially more pronounced effects on low-resource languages.
\end{itemize}

\paragraph{Evaluation Scope} Our evaluation spans a diverse but deliberately standardized set of NLP tasks commonly used in prior work on transfer language selection. While this does not cover all NLP tasks, it enables comparison and replication across studies. However, since the effects of language representations have been shown to be task-specific, the proposed representations are not guaranteed to be applicable to other tasks not studied here. Our results further demonstrate variability in performance even within the same tasks (such as between XNLI and SIB200), likely originating from other factors such as data domain, choice of model, language coverage, etc. Moreover, we focus on the application of language distances on choosing transfer languages using \textsc{LangRank} only; the utility of our language representations on other frameworks and/or applications remains unexplored.

\paragraph{Distance Measures}
While our work demonstrates the strength of distances from new language representations, these singular numerical distances, even in a focused direction, cannot fully capture the complexity in linguistic relationships. Furthermore, the task-agnostic composite distance we present should not be considered as universally effective. More complex, non-linear models, adapted to specific tasks, could potentially yield further gains, which we leave for future work.

To mitigate these issues and promote accessibility, we release our full codebase. Furthermore, while the speaker distributions cannot be released due to data licensing, we publicly release our Hyperbolic genetic embeddings and Latent Island typological representations to encourage more principled investigations into linguistic distance.

\section*{Ethics Statement}
The intention of this study is to enhance the representations of the world's languages, with the ultimate aim of improving cross-lingual performance, while promoting equity and inclusivity, in language technologies.

No personally identifiable or sensitive data was used in this study. However, our work relies on established linguistic knowledge bases and datasets, and we acknowledge that our work is subject to any biases or inaccuracies in these sources, which may under-represent low-resource languages or certain speaker communities.

We further recognize that our proposed methods may be computationally intensive, which can create barriers for researchers with limited computational resources. To promote accessibility and reproducibility, we release our code and language representation data where possible, including a limited subset of the speaker data under Ethnologue's Fair Use Guidelines.

\section*{Acknowledgments}
We thank Mason Shipton, Jun Bin Cheng and Junghyun Min for their feedback and exploratory work. This work was supported by the Fields Undergraduate Summer Research Program from the Fields Institute for Research in Mathematical Sciences (University of Toronto), and by the Undergraduate Summer Research Program from the Department of Computer Science at the University of Toronto. We also thank the anonymous reviewers for their constructive feedback.

\bibliography{EACL26Main}
\bibliographystyle{acl_natbib}

\appendix\label{app:appendix}

\section{Language Coverage of Representations}

We report the number of languages covered by our language representations in Table \ref{tab:language_coverage}.

\begin{table}[h!]
\centering
\begin{tabular}{l c}
\hline
\textbf{Representation} & \textbf{Number of languages} \\
\hline
Speaker & 6695 \\
Hyperbolic & 7836\\
Islands & 4555 \\
\hline
\end{tabular}
\caption{Number of languages with data per representation.}
\label{tab:language_coverage}
\end{table}

Although \uriel{+} nominally enables distance computations for 8171 languages, coverage within each modality varies, as the underlying data sources contain information for only subsets of languages. Our proposed representations are subject to similar limitations, namely being constrained by the language coverage of Ethnologue, Glottolog, and \uriel{+}. The hyperbolic embeddings represent families, languages, and dialects, totaling 26223 entities. Moreover, the combined breadth of these resources remains considerable, underscoring their utility in cross-lingual transfer particularly for less-resourced languages.

\section{Geographic Distance Metric Derivations}\label{app:geographic}
Here, we prove the normalization property of the geographic distance we discuss in Section \ref{subsec:geography}. Denote the Wasserstein-1 distance by $W_1$. We know that for any two languages $P,Q$ we have $W_1(P,Q) \leq D_{\max}$ because we can always design a transport plan $\pi$ such that $$\sum_{i=1}^{r}\sum_{j=1}^{n}\pi_{ij}c(y_i,z_j) \leq D_{\max}.$$
The details of this plan $\pi$ are as follows. For every $(i,j)$ pairing, we set $\pi_{ij} = q_i \cdot v_j$. We first check that this is a valid transport plan.

\begin{enumerate}
    \item It is clear that for all $i,j$, $q_i,v_j \geq 0$, $\pi_{ij} \geq 0$. 
    \item For any $i$, we see that $\displaystyle \sum_{j = 1}^n \pi_{ij} = \sum_{j = 1}^n (q_i \cdot v_j) = q_i \sum_{j = 1}^n v_j = q_i \cdot 1 = q_i$.
    \item For any $j$, we see that $\displaystyle \sum_{i = 1}^r \pi_{ij} = \sum_{i = 1}^r (v_j \cdot q_i) = v_j \sum_{i = 1}^r q_i = v_j \cdot 1 = v_j$.
\end{enumerate}

Hence, this is a valid plan. Then, we know that for any two points on earth $y,z$, that $d_g(y,z) = c(y,z) \leq D_{\max}$. Therefore, plugging this inequality into the above summation using the aforementioned transport plan gives us that

\begin{align*}
    &\sum_{i=1}^{r}\sum_{j=1}^{n}\pi_{ij}c(y_i,z_j) \\
    &\leq \sum_{i=1}^{r}\sum_{j=1}^{n}\pi_{ij} D_{\max} \\
    &= \sum_{i=1}^{r}\sum_{j=1}^{n}(q_i \cdot v_j)  D_{\max} \\
    &= \sum_{i=1}^{r}\sum_{j=1}^{n}(q_i \cdot v_j)  D_{\max} \\
    &= D_{\max}\sum_{i=1}^{r}q_i\sum_{j=1}^{n} v_j \\
    &= D_{\max}
\end{align*}

Now, from the definition of Wasserstein-1 distance, we know that 
$$W_1(P,Q) \leq \sum_{i=1}^{r}\sum_{j=1}^{n}\pi_{ij}c(y_i,z_j) \leq D_{\max},$$
and this statement is proved. In addition, normalizing based on antipodal distance is also the technique implemented by \uriel{+}, which gives credence to this normalization technique.

\section{Genetic Embedding: Geometry \& Optimization Details}\label{app:genetic}

This appendix contains the implementation details that were omitted from the main body but are necessary to reproduce the genetic embeddings in each geometry.

\subsection{Data Preparation}

We store the Glottolog genealogy as a directed adjacency list, constructed by parsing Glottolog’s Newick representation. The converter supports an optional dialect-pruning step: subtrees containing no language-level nodes are removed, yielding a graph in which languages have no outgoing edges and thus appear as leaves. Including dialectic information during the embedding process increases the parent language's centrality in hyperbolic space, which can affect pairwise genetic distances.

\subsection{Poincaré Ball Model}
We work in the open unit ball \(\mathcal{B}^d=\{\mathbf{x}\in\mathbb{R}^d:\|\mathbf{x}\|_2<1\}\) endowed with the Riemannian metric
\[
g_{\mathbf{x}}=\left(\frac{2}{1-\|\mathbf{x}\|_2^2}\right)^2 I_d.
\]
Translations use Möbius addition
\[
\mathbf{u}\oplus \mathbf{v}=
\frac{(1+2\langle \mathbf{u},\mathbf{v}\rangle+\|\mathbf{v}\|_2^2)\mathbf{u} + (1-\|\mathbf{u}\|_2^2)\mathbf{v}}
     {1+2\langle \mathbf{u},\mathbf{v}\rangle+\|\mathbf{u}\|_2^2\|\mathbf{v}\|_2^2},
\]
with the denominator clamped to \(\ge \epsilon\).
The optimization uses Riemannian stochastic gradient descent. Given a Euclidean gradient \(g_e\), it is first converted to a Riemannian gradient in the tangent space of \(\mathbf{x}\) by scaling:
\[
g_r = \frac{(1-\|\mathbf{x}\|_2^2)^2}{4} g_e.
\]
The update is then performed by moving along the geodesic in the direction of \(-g_r\):
\[
\mathbf{x}_{t+1} = \mathbf{x}_t \oplus \left(\tanh\left(\frac{\eta \lambda_{\mathbf{x}_t}\|g_r\|_2}{2}\right)\frac{-g_r}{\|g_r\|_2}\right),
\]
where \(\eta\) is the learning rate. After the update, if a point \(\mathbf{y}\) lands outside the unit ball due to numerical instability, it is projected back to the boundary by rescaling: \(\mathbf{y} \leftarrow \mathbf{y} \frac{1-\epsilon}{\|\mathbf{y}\|_2}\).
For the geodesic distance (defined in the main body), the argument of \(\cosh^{-1}(\cdot)\) is clamped to \(\ge 1+\epsilon\) for numerical stability.

\subsection{Hyperboloid Model}
We embed in \[\mathcal{H}^d=\{\mathbf{x}\in\mathbb{R}^{d+1}:\langle \mathbf{x},\mathbf{x}\rangle_{L}=-1,x_0>0\}\] with Lorentzian inner product \[\langle \mathbf{x},\mathbf{y}\rangle_{L}=-x_0y_0+\sum_{i=1}^d x_i y_i.\]
For the hyperbolic distance (defined in the main body), we clamp \(-\langle \mathbf{u},\mathbf{v}\rangle_{L}\) to \(\ge 1+\epsilon\). Optimization in the hyperboloid model is performed by applying the following update steps for a point \(\mathbf{x}\) with a corresponding Euclidean gradient \(g_e\):
\begin{enumerate}
    \item {Gradient Projection:} The Euclidean gradient \(g_e\) is projected onto the tangent space at \(\mathbf{x}\) to obtain the Riemannian gradient \(g_r\). Let \(g_e^{L}\) be the gradient with its time-like coordinate negated. Then,
    \[ g_r = g_e^{L} + \langle \mathbf{x}, g_e^{L} \rangle_{L} \mathbf{x}. \]
    \item {Gradient Clipping:} The norm of the Riemannian gradient is clipped to a maximum value of \(c_g\): 
    \[ g_r \leftarrow g_r \cdot \min\left(1, \frac{c_g}{\|g_r\|_{L}}\right). \]
    \item {Exponential Map:} The point is updated by moving along the geodesic. The tangent vector for the update is \(\mathbf{u} = -\eta g_r\), where \(\eta\) is the learning rate. This produces an intermediate point, \(\tilde{\mathbf{x}}\):
    \[ \tilde{\mathbf{x}} = \cosh(\|\mathbf{u}\|_{L})\mathbf{x}_t + \sinh(\|\mathbf{u}\|_{L})\frac{\mathbf{u}}{\|\mathbf{u}\|_{L}}. \]
    \item {Manifold Projection:} As a final safeguard, the intermediate point \(\tilde{\mathbf{x}}\) is projected back to the hyperboloid to yield the final updated point \(\mathbf{x}_{t+1}\). This step also prevents numerical overflow by clipping the norm of the spatial components of \(\tilde{\mathbf{x}}\) (denoted \(\tilde{\mathbf{x}}_{1:}\)) to a maximum of \(c_s\):
    \[ \mathbf{x}_{t+1} = \Big[\sqrt{\|\mathbf{x}'_{1:}\|_2^2+1}, \mathbf{x}'_{1:}\Big]\]\[ \text{where }  \mathbf{x}'_{1:} = \tilde{\mathbf{x}}_{1:}\cdot \min\left(1,\frac{c_s}{\|\tilde{\mathbf{x}}_{1:}\|_2}\right). \]
\end{enumerate}
The clipping thresholds \(c_g\) and \(c_s\) are hyperparameters. 

\begin{table}[ht!]
\centering
\begin{tabular}{llrr}
\toprule
\textbf{Geometry} & \textbf{Dim} & \multicolumn{1}{c}{\textbf{MR}} & \multicolumn{1}{c}{\textbf{MAP}} \\
\midrule
\multirow{4}{*}{Hyperboloid} & 2 & 6.3329 & 0.6743 \\
 & 5 & 2.5227 & 0.8723 \\
 & 10 & 1.3674 & 0.9513 \\
 & 50 & 1.2518 & 0.9581 \\
\cmidrule(lr){2-4}
\multirow{4}{*}{Poincar\'e} & 2 & 6.9936 & 0.5969 \\
 & 5 & 2.1246 & 0.8601 \\
 & 10 & 2.0591 & 0.8633 \\
 & 50 & 2.1478 & 0.8463 \\
\cmidrule(lr){2-4}
\multirow{4}{*}{Euclidean} & 2 & 274.0730 & 0.1910 \\
 & 5 & 147.7106 & 0.3043 \\
 & 10 & 56.3716 & 0.4286 \\
 & 50 & 3.3975 & 0.7180 \\
\bottomrule
\end{tabular}
\caption{Reconstruction performance on the ancestor retrieval task. We report Mean Rank (MR) and Mean Average Precision (MAP) for each geometry across varying embedding dimensions (Dim). }
\label{tab:recon_results}
\end{table}

\subsection{Reconstruction Metrics and Results}\label{app:metrics}

To evaluate how well the learned embeddings capture the original hierarchical structure, we perform a link prediction task focused on ancestor-descendant relationships. For each node \(u\) in the graph \(V\), we rank all other nodes \(v \in V \setminus \{u\}\) based on their geometric distance \(d(u, v)\) in ascending order. We treat the set of true ancestors of \(u\), denoted \(\mathcal{A}(u)\), as the positive items to be retrieved. From this ranking, we compute two retrieval metrics: Mean Rank (MR) and Mean Average Precision (MAP).

\paragraph{Mean Rank (MR)}
This metric measures the average rank of a true ancestor. For each descendant-ancestor pair \((u, a)\) where \(a \in \mathcal{A}(u)\), we compute the rank of \(a\) in the distance-sorted list of nodes relative to \(u\). A lower MR indicates better performance, as it means true ancestors are, on average, found closer to their descendants in the embedding space. The rank is formally defined as:
$
\text{rank}(a, u) = 1 + \left| \{ v \in V \setminus (\mathcal{A}(u) \cup \{u\}) : d(u, v) < d(u, a) \} \right|.
$
The final MR is the average of these ranks over all true descendant-ancestor pairs in the graph.

\paragraph{Mean Average Precision (MAP)}
MAP provides a more comprehensive measure of ranking quality by rewarding models that place many true ancestors early in the ranked list. For each node \(u\), we first compute its Average Precision (AP), which is the average of precision values at each rank \(k\) that contains a true ancestor:
\[
\text{AP}(u) = \frac{\sum_{k=1}^{|V|-1} P(k) \times \mathbb{I}(v_k \in \mathcal{A}(u))}{|\mathcal{A}(u)|},
\]
where \(v_k\) is the node at rank \(k\), \(P(k)\) is the precision at rank \(k\) (i.e., the fraction of true ancestors in the top \(k\) results), and \(\mathbb{I}(\cdot)\) is the indicator function. The final MAP score is the mean of these AP scores over all nodes in the graph. A higher MAP score indicates better performance.

\begin{figure*}[t!]
    \centering
    \includegraphics[width=1\linewidth]{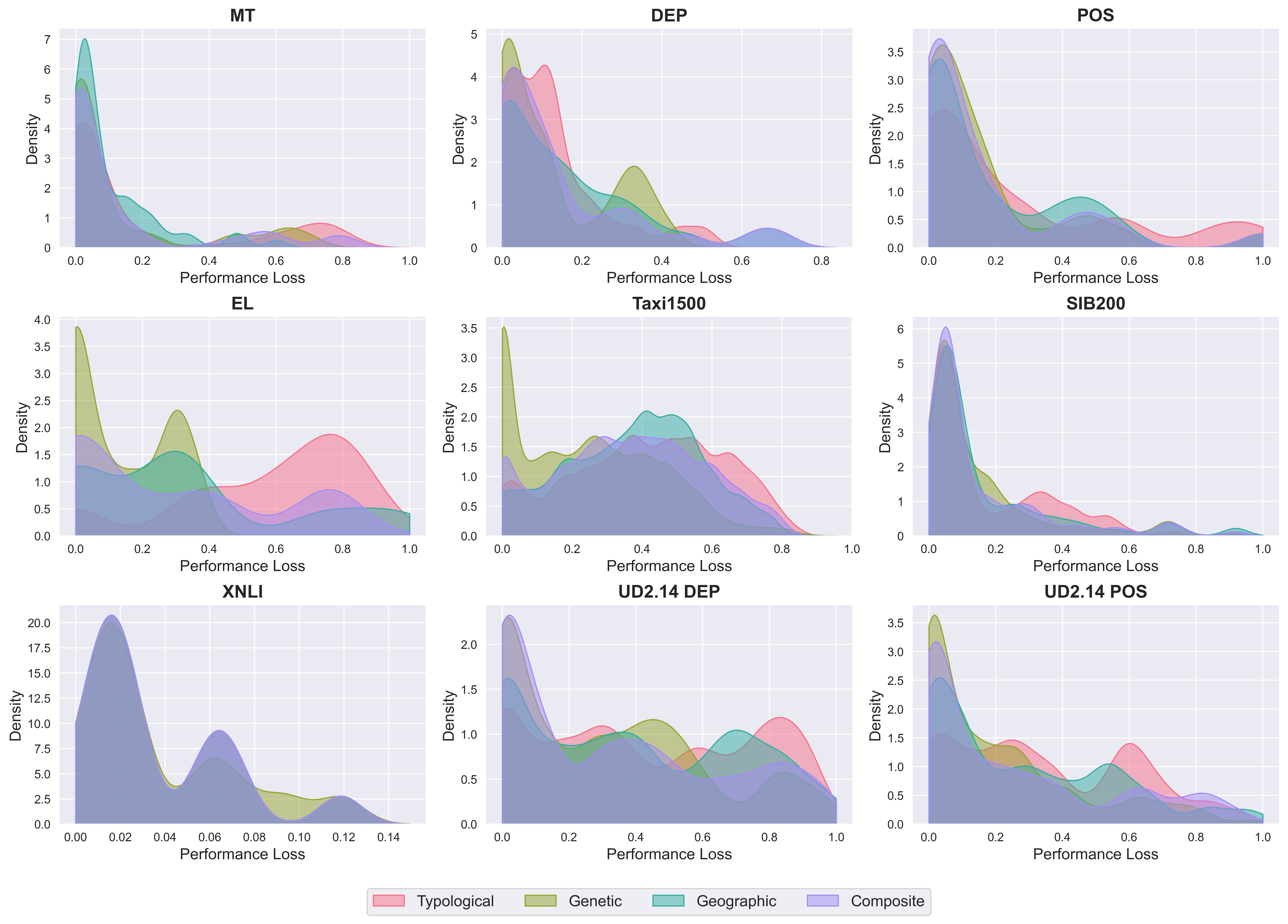}
    \caption{Kernel density estimates of performance loss for URIEL+ distances across tasks. The composite distance yields more peaked distributions.}
    \label{fig:composite-1}
\end{figure*}

\begin{figure*}[t!]
    \centering
    \includegraphics[width=1\linewidth]{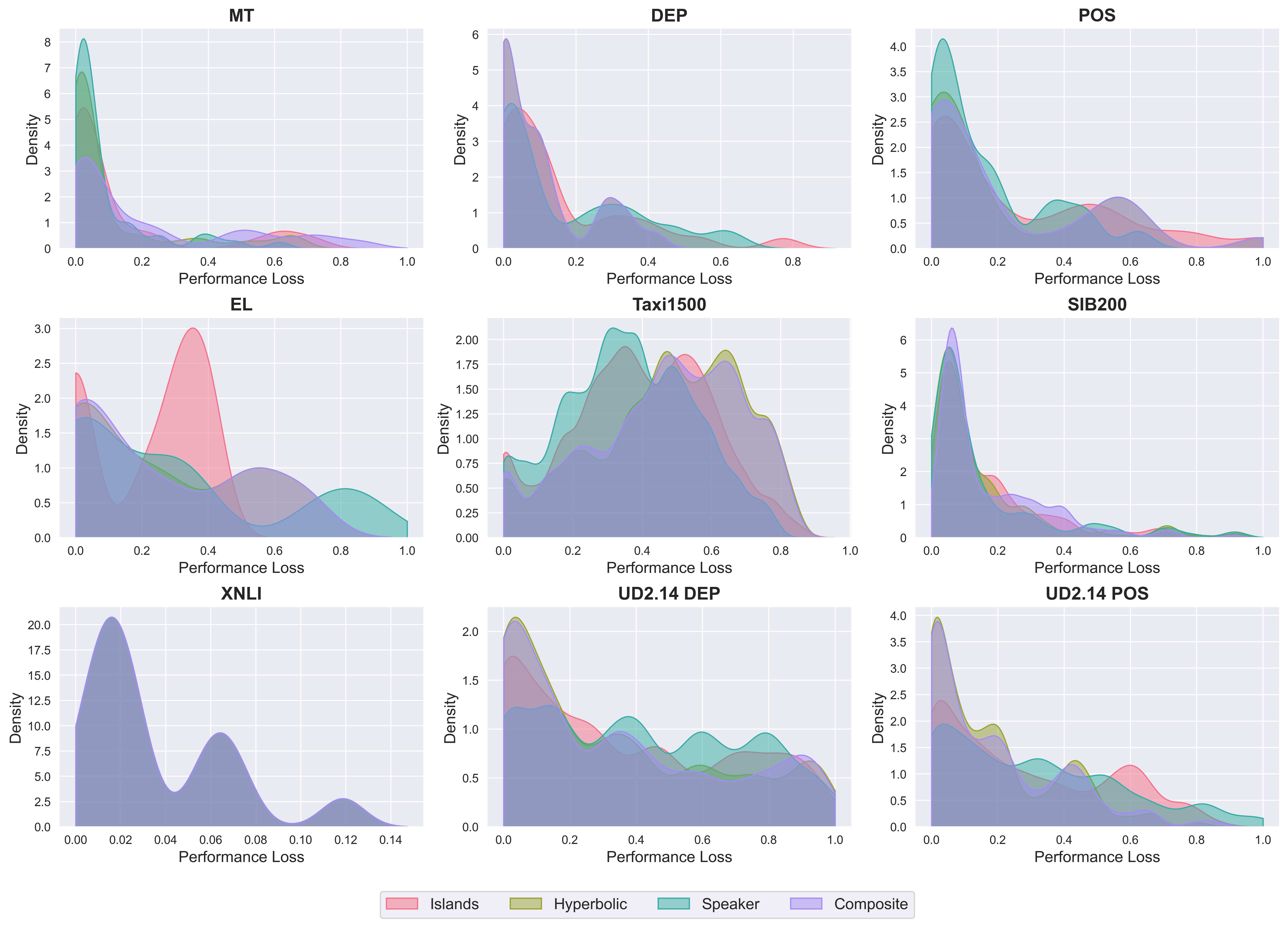}
    \caption{Kernel density estimates of performance loss for our modality-matched distances across tasks. Similarly, the composite distance yields more peaked distributions.}
    \label{fig:composite-2}
\end{figure*}

\paragraph{Results}
The performance of our genetic embedding algorithm across different geometries and dimensions is summarized in Table~\ref{tab:recon_results}. The results clearly show that hyperbolic geometries (Hyperboloid and Poincaré) significantly outperform Euclidean geometry, especially at lower dimensions. The Hyperboloid model consistently achieves the best scores, demonstrating its effectiveness in capturing the hierarchical relationships of the data. Hence, we select the Hyperboloid model.

\section{Implementation Details for Latent Tree Models.}
\label{app:latent}

We employ a modified Bayesian Information Criterion (BIC) defined as $2k^2\log(n)-2\mathbb{L}$, where $k$ denotes the number of parameters, $\mathbb{L}$ is the log-likelihood, and $n$ is the number of samples. This modified criterion, which penalizes the number of parameters quadratically, more strongly discourages models with a large number of free parameters compared to the traditional linear penalty. In our greedy clustering context, this helps prevent the algorithm from forming many small, fragmented clusters, instead favoring more balanced and structurally coherent feature islands. When computing the BIC values for two clusters, there is a higher penalty for imbalanced cluster sizes.

To learn a latent variable for a subset of features, we run the Expectation–Maximization algorithm with five restarts with random initializations to mitigate the risk of convergence to local optima.The resulting model yields 325 feature clusters, each associated with a latent variable. Cluster sizes range from 1 to 11. To assess effectively in grouping correlated features, we compute the absolute Pearson correlation among features in each cluster to measure intra-cluster association strength. For clusters of size three or larger, the average absolute correlation is $0.623$, indicating that features grouped together tend to be strongly correlated. Clusters of size $\leq 2$ are excluded from this analysis.

\section{Analysis and Extensions of Composite Distances}\label{app:composite}
\subsection{Distributional Analysis}\label{app:composite-distribution}
Table \ref{tab:composite} demonstrates that a single, task-agnostic composite score, averaging over our modality-matched language distances, yields performance gains over using \langrank{} with multiple \uriel{+} distances. While this presents task-agnostic composite distances as a robust alternative where training task-specific models is not feasible, we aim to demonstrate its stability over its individual constituents as well. To study the behavior of task-agnostic composite distances, we further examine the distributions of performance losses from two composite distances: (1) averaging over \uriel{+} distances, and (2) averaging over our proposed modality-matched distances, and compare them against its constituent distances.

\paragraph{\uriel{+} Distances} Figure \ref{fig:composite-1} shows kernel density estimates of performance loss for URIEL+ typological, genetic, geographic, and composite (a simple average of the three) distances across tasks. Individual modalities often exhibit polarized behavior: sharp peaks near zero for some tasks (e.g. typological distance on POS), but heavier tails (e.g. typological distance on EL) or secondary modes (e.g. typological distance on Taxi1500) for others, reflecting task-specific modality relevance. The URIEL+ composite distance consistently produces more central distributions, smoothing extreme behaviors across individual modalities and reducing variance in performance loss across tasks.

\paragraph{Modality-Matched Distances} Figure \ref{fig:composite-2} presents the same analysis for our proposed distances and their composite. As in the URIEL+ setting, task-specialized distances can closely match the ideal distribution for particular tasks, but may exhibit heavier tails elsewhere. In the majority of tasks, composite distance yields distributions with mass concentrated near low loss while avoiding pronounced secondary modes.

Across both settings, these task-agnostic composite distances do not uniformly minimize loss. Instead, they more consistently approximate the ideal distribution, with higher mass nearer to zero with moderated tails, across diverse tasks. This reinforces our finding that, while the effectiveness of individual modalities are task-dependent, a single composite score, even one which is task-agnostic, can remain robust across tasks. Moreover, this suggests that task-adapted composite distances may yield further task-specific gains.

\subsection{Task-Specific Weights}\label{app:composite-weights}

Although one can learn the weights in a number of different ways, we present one simple method using the performance losses from our \langrank{} evaluation framework. If $l_p \in [0, 1]$ is some performance loss (e.g. accuracy, F1, or RMSE if it is known to be in the unit interval), then $1-l_p$ gives a measure of the quality of performance on a given task. In this case, one can use each of the modality distances $d^m$ as covariates to predict $l_p$, say via a linear regression. Upon obtaining the coefficient estimates, one can take the coefficients into $[0, 1]$. Common options include transforming each coefficient estimate by the logistic function (or ReLU) and then normalizing. 

\section{Downstream Task Setup Details}\label{app:downstream}

Our objective is to design an evaluation (tasks, evaluation metric) which is closely aligned with actual applications of language distances in cross-lingual transfer. In particular, the usage of language distances on choosing source languages has been widely studied (see Section \ref{sec:related-work}). We therefore focus on applying new language representations to \langrank{} \cite{lin-etal-2019-choosing}, a commonly used framework for choosing source languages for a given NLP task.

We mostly replicate \citet{lin-etal-2019-choosing} and \citet{khan-etal-2025-uriel}'s pipeline for evaluating distances using \langrank{}. This process involves first collecting, for a given NLP task (e.g. Taxi1500 topic classification) and model (e.g. mBERT), a dataset of performance scores for each target and source language pair. Next, during evaluation, we perform leave-one-language-out cross-validation by holding out scores for each target language, training a LightGBM ranker on the remaining data (additionally holding out 10\% of data as a validation set), and evaluating the ranker on how well it picks source languages for the held-out target language.

\subsection{Experimental Datasets}
With \langrank{}, we evaluate the utility of distances by applying them to a diverse set of nine sub-tasks. For the first four (DEP, EL, MT, POS) we re-use the performance datasets provided by \citet{lin-etal-2019-choosing}. We additionally derived performance datasets for each new task studied:
\begin{itemize}
\item \textbf{Taxi1500:} Due to the infeasibility of training models for each language covered by Taxi1500, we train 33 mBERT \cite{devlin-etal-2019-bert} models according to the languages in Taxi1500 which are defined as high- or medium-resource in \uriel{+}, evaluating each model's performance on the 799 languages whose data is publicly available and contains >900 examples. 
\item \textbf{SIB200 \& XNLI:} We train one model for each language, (in SIB200, rejecting 37 languages where the model did not converge), and finally evaluating each model on the test splits of all other languages. 
\item \textbf{UD v2.14:} We replicate the setup from \citet{blaschke-etal-2025-analyzing}, and simply evaluate the test split of each language on each of the 70 UDPipe2 \cite{straka-2018-udpipe} models, averaging scores over treebanks within the same language.
\end{itemize}

For each task, we use the same train-validation-test splits as published.

\subsection{Evaluating Distances}\label{app:downstream-metric}

After collecting datasets, we run \langrank{} and ablate on, for each modality, training with distances computed from the \uriel{+} representation versus our new representation. We measure its performance with the performance loss metric $l$, which are averaged across folds, to showcase the real-world implications of our \langrank{} experiments. Here, we define performance loss $l_i$ for the fold associated with holding out target language $i$ as:
\[
l_i = \frac{(\max_j s_{ij}) - s_{ik}}{\max_j s_{ij}}
\]

where $k$ is the top-1 language chosen by \langrank{}, and score $s_{ij}$ refers to the model performance on the given NLP task when transferring to language $i$ from language $j$. Simply put, given a particular model and a particular NLP task, performance loss $l$ measures the relative difference in model performance between transferring using \langrank{}'s chosen language and the optimal language.

In particular, we choose to consider only the top-1 chosen language due to the observation that practitioners often choose only the top-1 language (as opposed to, e.g. trying all top-3 languages) to perform cross-lingual transfer. This decision therefore aligns with our underlying objective of designing a realistic evaluation setup.

To isolate the effect of individual distance representations while accounting for variability across cross-validation folds, we conduct an ablation study using a linear mixed-effects model. We model the performance score as a function of the typological, geographic, and genetic representations, treated as categorical fixed effects, with a random intercept for each fold. Formally, for each evaluation instance $i$, we fit:
\begin{align*}
\text{score}_{i}&=\beta_0 + \beta_{\mathrm{typ}}^{(k_i)} + \beta_{\mathrm{geo}}^{(g_i)}
+ \beta_{\mathrm{gen}}^{(h_i)} + u_{f_i}, \\
u_{f_i} &\sim \mathcal{N}(0, \sigma_f^2), \qquad
\epsilon_i \sim \mathcal{N}(0, \sigma^2),
\end{align*}
where $k_i$, $g_i$, and $h_i$ index the typological, geographic, and genetic representations used for instance $i$, respectively, and $u_{f_i}$ is a random intercept associated with cross-validation fold $f_i$.

\begin{table}[!t]
\centering
\begingroup
\footnotesize
\setlength{\tabcolsep}{2pt}
\setlength{\fboxsep}{2pt}
\begin{tabular}{l l c c}
\toprule
\textbf{Modality} & \textbf{Representation} & LLaMA-3.1 & mBERT \\
\midrule
\multicolumn{2}{c}{Baseline:}   & \baselinecell{40.8}{0.6} & \baselinecell{38.1}{0.5} \\
\midrule
\multirow{2}{*}{Typ} & Laplacian & \celldata{-1.2}{0.5}{-2.9}{0} & \celldata{+0.4}{0.3}{0.9}{1} \\
& Islands    & \celldata{-1.2}{0.5}{-2.9}{0}   & \celldata{-0.9}{0.3}{-2.2}{0}\\
Geo & Speaker & \celldata{0.6}{0.4}{1.5}{1}  & \celldata{-2.1}{0.2}{-5.1}{0} \\
Gen & Hyperbolic    & \celldata{1.0}{0.4}{2.5}{0}  & \celldata{2.7}{0.2}{6.6}{0}  \\
\bottomrule
\end{tabular}
\endgroup

\caption{Taxi1500 topic classification using LLaMA-3.1-8B and mBERT. Regression coefficients measure baseline performance loss (using URIEL+ distances) and changes in loss when substituting alternative distance representations. Values are reported as mean ± standard error; \textbf{bold} indicates $p < 0.05$. Lower values indicate better transfer language selection.}
\label{tab:llama-results}

\end{table}

\subsection{Taxi1500 with LLaMA-3.1}\label{app:downstream-llama}
To address concerns regarding the age of models in our main evaluation, we additionally re-ran the Taxi1500 topic classification experiment using LLaMA-3.1-8B \cite{grattafiori2024llama3herdmodels}, a contemporary large language model with strong multilingual capabilities. We replicate the experiment in Section \ref{sec:evaluation}, differing only in the underlying model.

Table \ref{tab:llama-results} reports regression coefficients measuring baseline performance loss and changes in loss (in percentage points) when substituting URIEL+ distances with alternative representations. For reference, we also include the corresponding results for mBERT from Table \ref{tab:results}. Across both models, baseline performance losses are comparable, and statistically significant effects remain consistent: representations that significantly reduce (or increase) loss under mBERT do so under LLaMA-3.1 as well. For example, the typological islands representation significantly reduces loss in both settings, while hyperbolic genetic distances significantly increase loss in both models.

These results suggest that the effects of language distance representations are not tied to a specific underlying model, and that the task-dependent patterns identified in our main evaluation persist under modern large language models.

\subsection{Computational Setup}
\paragraph{Hyperparameters.}
We adopt the following hyperparameters for the LightGBM ranker:
\begin{itemize}
\item \texttt{Early stopping rounds}: 25
\item \texttt{Learning rate}: 0.1
\item \texttt{Min data in leaf}: 10
\item \texttt{Lambda L2}: 0.2
\end{itemize}

These hyperparameters were obtained by performing a grid search, and measuring the task-averaged \langrank{} performance when using baseline \uriel{+} distances.

For training transfer models in tasks Taxi1500 and SIB200, since per-language data is relatively scarce ($\sim$1k examples), we employ the following training arguments:
\begin{itemize}
\item \texttt{Num train epochs}: 10
\item \texttt{Learning rate}: 1e-5
\item \texttt{Batch size}: 16
\item \texttt{Eval steps}: 20
\item \texttt{Early stopping patience}: 5
\item \texttt{Weight decay}: 0.01
\item \texttt{Warmup ratio}: 0.1
\end{itemize}

For XNLI, we replicate the setup from \citet{philippy-etal-2023-identifying}, with the following training arguments:
\begin{itemize}
\item \texttt{Num train epochs}: 3
\item \texttt{Learning rate}: 2e-5
\item \texttt{Batch size}: 32
\end{itemize}

\begin{table*}[h]
\centering
\begin{tabular}{ll}
\toprule
\textbf{Artifact} & \textbf{License} \\
\midrule
\multicolumn{2}{l}{\textit{Packages}} \\
\uriel{+} \citep{khan-etal-2025-uriel} & CC BY-SA 4.0 \\
\langrank{} \citep{lin-etal-2019-choosing} & BSD 3-Clause \\
\midrule
\multicolumn{2}{l}{\textit{Datasets}} \\
Glottolog (v5.2) \citep{Glottolog} & CC BY 4.0 \\
Ethnologue (Edition 28) \citep{eberhard2025ethnologue28} & Proprietary (Licensed under SIL International) \\
Taxi1500 (v3) \citep{ma-etal-2025-taxi1500} & Apache 2.0 \\
XNLI \citep{conneau-etal-2018-xnli} & CC BY-NC 4.0 \\
SIB200 \citep{adelani-etal-2024-sib} & CC BY-SA 4.0 \\
UD (v2.14) \citep{11234/1-5502} & Various \\
\midrule
\multicolumn{2}{l}{\textit{Models}} \\
Multilingual BERT cased \citep{devlin-etal-2019-bert} & Apache 2.0 \\
XLM-RoBERTa-base \citep{conneau-etal-2018-xnli} & MIT \\
UDPipe v2.12 \citep{straka-2018-udpipe} & MPL 2.0 \\
LLaMA-3.1-8B \citep{grattafiori2024llama3herdmodels} & llama3.1 \\
\bottomrule
\end{tabular}
\caption{Artifacts used in this study, and their licenses.}
\label{tab:artifacts}
\end{table*}

\paragraph{Computing Infrastructure}
Model training and evaluation for collecting \langrank{} experimental datasets were conducted on a single NVIDIA A100, requiring around 100 compute hours.

All actual \langrank{} experiments were performed on an Apple M1 Pro over 8 hours.

\section{Licenses for Artifacts Used}
\label{app:licenses}
The artifacts employed in this study, along with their respective licenses, are listed in Table \ref{tab:artifacts}.

All artifacts and datasets were used for the purpose of studying language representations, and were handled in accordance with their respective licenses.

\section{Use of Generative AI}
\label{app:genai}
Generative AI was employed only in a limited capacity: to assist in organizing and clarifying text, and to suggest code auto-completions during the implementation of experiments.

\end{document}